\documentclass[letterpaper, 10 pt, conference]{ieeeconf}
\IEEEoverridecommandlockouts    % This command is only needed if
\let\labelindent\relax
\usepackage{enumitem}

\usepackage{graphics}           
\usepackage{times}              
\usepackage{amsmath}            
\usepackage{amssymb}            
\usepackage{graphicx}
\usepackage{algorithm}
\usepackage[noend]{algpseudocode}
\usepackage{booktabs}
\usepackage{color}
\usepackage{xcolor}
\definecolor{instructioncolor}{rgb}{.5,.5,.5}

\usepackage[font=small]{caption}

\def\secref#1{Sec.~\ref{#1}}
\def\figref#1{Fig.~\ref{#1}}
\def\tabref#1{Tab.~\ref{#1}}
\def\eqref#1{Eq.~(\ref{#1})}

\makeatletter
\usepackage{xspace}
\DeclareRobustCommand\onedot{\futurelet\@let@token\@onedot}
\def\@onedot{\ifx\@let@token.\else.\null\fi\xspace}
 
\def\ie{i.e\onedot}

\def\etal{{et al}\onedot}
\makeatother

\def\etalcite#1{\etal~\cite{#1}}

\usepackage{array}
\newcolumntype{L}[1]{>{\raggedright\let\newline\\\arraybackslash\hspace{0pt}}m{#1}}
\newcolumntype{C}[1]{>{\centering\let\newline\\\arraybackslash\hspace{0pt}}m{#1}}
\newcolumntype{R}[1]{>{\raggedleft\let\newline\\\arraybackslash\hspace{0pt}}m{#1}}

\renewcommand{\b}[1]{\mbox{\boldmath$#1$}}

\newcommand{\bt}{\b t}

\renewcommand{\and}{\hspace{1cm}}

\title{\LARGE \bf On Domain-Specific Pre-Training \\ for Effective Semantic Perception in Agricultural Robotics}

\author{Gianmarco Roggiolani \and Federico Magistri \and Tiziano Guadagnino \and Jan Weyler\\[1.2mm] Giorgio Grisetti \and Cyrill Stachniss \and Jens Behley% <-this % stops a space
  \thanks{G.~Roggiolani. F.~Magistri, T.~Guadagnino, J.~Weyler, J.~Behley, and C.~Stachniss are with the University of Bonn, Germany. C.~Stachniss is additionally with the Department of Engineering Science at the University of Oxford, UK, and with the Lamarr Institute for Machine Learning and Artificial Intelligence, Germany. G.~Grisetti is with La Sapienza University of Rome, Italy.}%
  \thanks{This work has partially been funded 
  by the Deutsche Forschungsgemeinschaft (DFG, German Research Foundation) under Germany's Excellence Strategy, EXC-2070 -- 390732324 -- PhenoRob, and
    by the Deutsche Forschungsgemeinschaft (DFG, German Research Foundation) under STA~1051/5-1 within the FOR 5351~(AID4Crops).
  }%
}

\begin{document}
\maketitle
\thispagestyle{empty}
\pagestyle{empty}

%%%%%%%%%%%%%%%%%%%%%%%%%%%%%%%%%%%%%%%%%%%%%%%%%%%%%%%%%%%%%%%%%%%%%%%%%%%%%%%%
\begin{abstract}
Agricultural robots have the prospect to enable more efficient and sustainable agricultural production of food, feed, and fiber. Perception of crops and weeds is a central component of agricultural robots that aim to monitor fields and assess the plants as well as their growth stage in an automatic manner. Semantic perception mostly relies on deep learning using supervised approaches, which require time and qualified workers to label fairly large amounts of data. In this paper, we look into the problem of reducing the amount of labels without compromising the final segmentation performance.
For robots operating in the field, pre-training networks in a supervised way is already a popular method to reduce the number of required labeled images. We investigate the possibility of pre-training in a self-supervised fashion using data from the target domain. To better exploit this data, we propose a set of domain-specific augmentation strategies. 
We evaluate our pre-training on semantic segmentation and leaf instance segmentation, two important tasks in our domain. 
The experimental results suggest that pre-training with domain-specific data paired with our data augmentation strategy leads to superior performance compared to commonly used pre-trainings. Furthermore, the pre-trained networks obtain similar performance to the fully supervised with less labeled data.
\end{abstract}

%%%%%%%%%%%%%%%%%%%%%%%%%%%%%%%%%%%%%%%%%%%%%%%%%%%%%%%%%%%%%%%%%%%%%%%%%%%%%%%%
\section{Introduction}
\label{sec:intro}

%%%%%%%%%%%%%%%%%%%
%% WHY: 
% First, answer the WHY question: Why is that relevant? Why should I be
% motivated to read the paper? Why should I care? (1 paragraph, 2-5 sentences)

Sustainable crop production is fundamental to meet the increasing request for food, fuel, and fiber. It, however, must become more effective to fulfill all demands. Furthermore, the lack of workers is a key challenge, which is even increased during the recent COVID pandemic. Robots are a crucial component to analyze and monitor plants in an automated way~\cite{marks2022icra}, followed by targeted fertilizing and/or protection~\cite{fiorani2013arpb}. 
Before targeted actions can be performed, the underlying perception problems need to be solved. Deep learning approaches improved the performance of these systems using large neural networks. Such networks for object detection or semantic segmentation can help the farmers to evaluate the status of the plants~\cite{halstead2021fps, magistri2020iros}, spot and locate weeds~\cite{weyler2021ral}, detect plant diseases \cite{goerlich2021drones}, and understand the growing conditions among different areas of the field~\cite{weyler2022ral}.

\begin{figure}[t]
  \centering
  \hspace{-0.2cm}\includegraphics[width=0.92\linewidth]{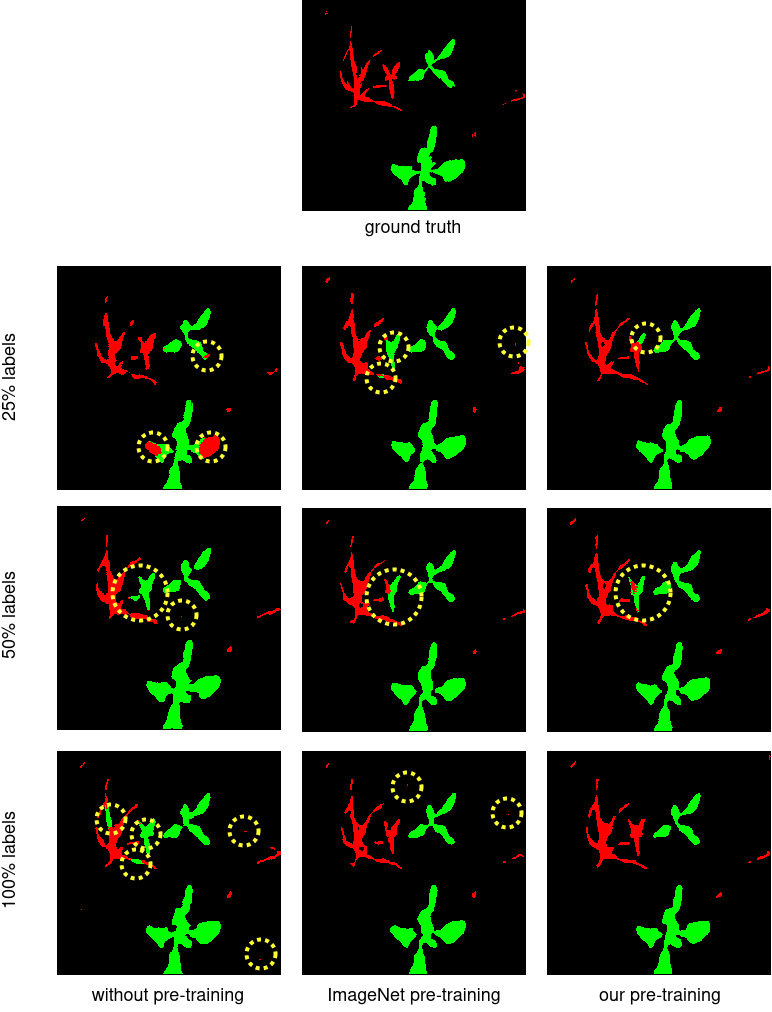}
  \caption{Results on semantic segmentation with different amounts of data and pre-trainings. We achieve better or comparable results with $\frac{1}{4}$ of the epochs and $\frac{1}{1000}$ of the images. The dotted circles highlight the errors in the results.}
  \label{fig:motivation}
  \vspace{-0.55cm}
\end{figure}

\begin{figure}[t]
	\hspace{0.1in}
	\includegraphics[width=\linewidth]{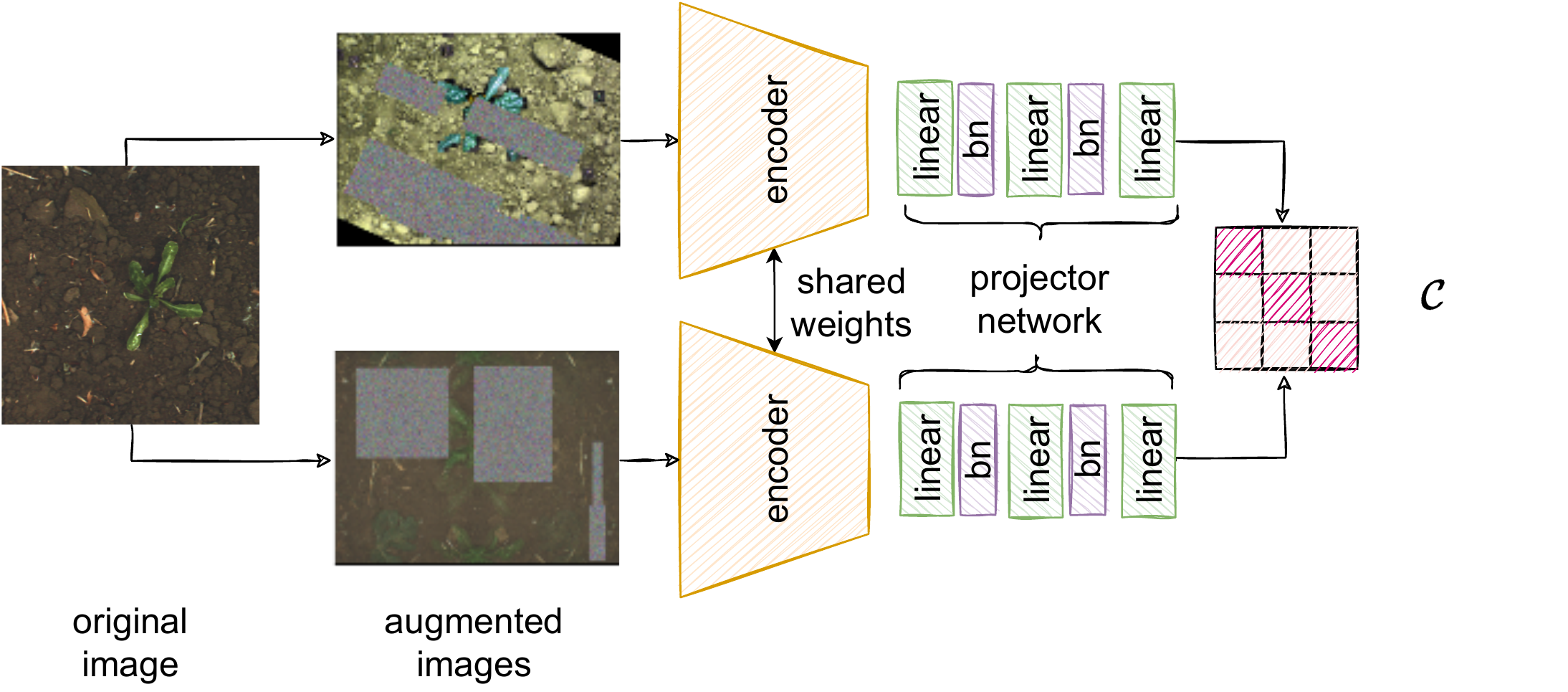}
	\caption{\label{fig:arch} For each image we build two augmentations to produce two embeddings. We fill a matrix $\mathcal{C}$ with their cross-correlation and then compute the loss, that forces $\mathcal{C}$ to be an identity.}
	\vspace{-0.5cm}
\end{figure}

Deep learning-based approaches, however, generally need a large amount of labeled data, which is hard to get because it needs time and specialized workers.
Some researchers use semi-supervised approaches to reduce the need for labeled images~\cite{lottes2017iros} or leverage background knowledge~\cite{milioto2018icra}. 
Recent work in self-supervised pre-training showed promising results, where we can pre-train a network without relying on supervision by labels. Pre-training on the ImageNet dataset~\cite{deng2009cvpr} is a common way to reduce the number of training samples and the training time for the network to converge. Several other methods use a contrastive loss and strong augmentation techniques~\cite{caron2021neurips, chen2020icml, grill2020neurips, he2020cvpr-mcfu, zbontar2021icml}. Embeddings produced in such a way are more robust to the augmentations applied. However, the applied data augmentations need to be selected carefully so as not to lose relevant features.
In this work, we study self-supervised representation learning to improve the perception of agricultural robots.  

The main contribution of this paper is providing a pre-training strategy for the plant domain, which will reduce the number of labeled images needed and a newly defined augmentation policy. We study existing augmentation and propose domain-specific ones to boost performance. Specifically, we target semantic and leaf instance segmentation and investigate how self-supervised pre-training on domain-specific data leads to better models that can learn with less labeled data.
In sum, we make three key claims:
(i)~pre-training on datasets of the same domain improves the performance of the downstream tasks,
(ii)~using domain-specific pre-training can further reduce the number of labeled images needed, and
(iii)~the augmentation policy needs to be domain-specific and take into account the order and design of the augmentations.
Our code is available at \texttt{https://github.com/ PRBonn/agri-pretraining}.
\begin{figure*}[t]
	\hspace{-0.15in}
	\vspace{-0.2in}
	\includegraphics[width=1.03\linewidth]{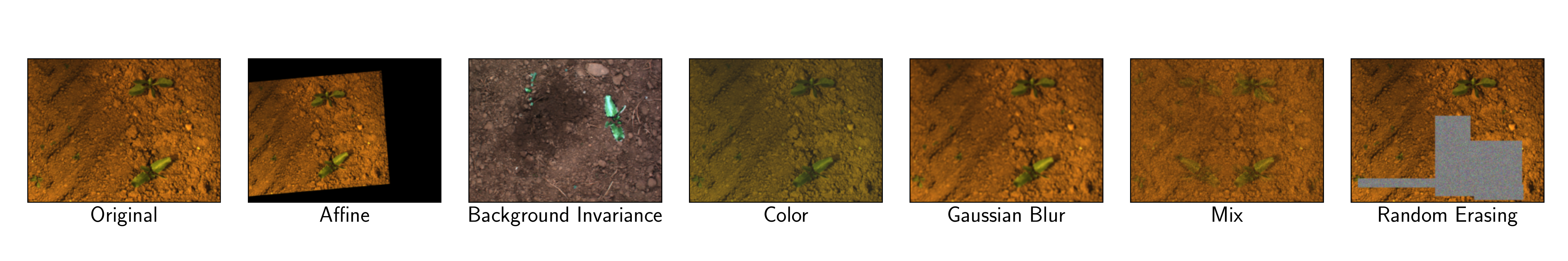}
	\caption{\label{fig:augmentations}We applied all of our augmentations to a single image to show one possible outcome for each one. In our pre-training strategy they are applied sequentially producing even more variations of the input.}
	\vspace{-0.35cm}
\end{figure*}
%%%%%%%%%%%%%%%%%%%%%%%%%%%%%%%%%%%%%%%%%%%%%%%%%%%%%%%%%%%%%%%%%%%%%%%%%%%%%%%%
\section{Related Work}
\label{sec:related}

% Discuss the main related work and cite around 15-25 papers in sum. 
% The related work section should be approx. 1 column long, assuming 
% a 6-page paper.  Structure the section in paragraphs, grouping the 
% papers, and describing the key approaches with 1-2 sentences. If 
% applicable, describe the key difference to your approach at the end 
% of each paragraph briefly. Avoid adding subsections, al least for a 
% conference paper.

Robotic applications in agriculture aim at improving field monitoring and interventions diminishing the use of agricultural chemical inputs and production costs~\cite{pretto2020ram,vysotska2019rsswsabstract}. 
Semantic segmentation and instance segmentation are two key steps for weed control and plant phenotyping. The majority of recent solutions for these tasks employ large convolutional neural networks~\cite{long2015cvpr-fcnf,lottes2018iros, pretto2020ram}, requiring large amounts of labeled data. Some of them exploit spatial information about the fields~\cite{lottes2018ral} or vegetation indexes~\cite{milioto2018icra}, while others focused on the architecture side, as Potena~\etalcite{potena2016ias} where they use two CNNs to detect the vegetation and then classify it, Weyler~\etalcite{weyler2021ral} which employs a Feature Pyramid Network to detect the instances, or in the works from Buzzy~\etalcite{buzzy2020sensors} and You~\etalcite{you2020computersinagriculture} that use deep neural networks.

Two main paths have been investigated to reduce the number of labeled images: domain adaptation and pre-training. In the domain adaptation setting, the network learns how to perform a task on a source domain and is expected to have a good performance on a different one. Approaches often use generative adversarial networks~\cite{liu2016nips,yoo2016eccv}. Their application on the plant domain shows already promising results for classification~\cite{gogoll2020iros}, object counting~\cite{ayalew2020eccv}, and object detection~\cite{hartley2021plants}. However, these methods require training the network on a source domain dataset for which we might need vast amounts of labeled images.

In contrast, pre-training aims at initializing the weights of a neural network, such that it needs fewer labels and converges faster. Erhan~\etalcite{erhan2010jmlr} shows that unsupervised pre-training guides the networks towards the minimum, from where supervised training can proceed faster and with less available data. In literature, pre-training on ImageNet is a common choice~\cite{razavian2014arxiv}, since it is big enough to provide a good initialization for various tasks and domains. 

Lately, self-supervised pre-training received increasing attention due to its promising results on several downstream tasks compared to supervised pre-training~\cite{chen2020icml, he2020cvpr-mcfu, chen2020arxiv, zbontar2021icml}. Especially, contrastive approaches firstly used large memory banks~\cite{wu2018cvpr} that were later replaced by a momentum encoder~\cite{he2020cvpr-mcfu} to reduce the required memory to store negative samples. In particular, Chen et al. \cite{chen2020icml} introduced a projection head for learning embeddings of positive and negative examples that showed superior performance and were later integrated into momentum contrast~\cite{chen2020arxiv}. They also investigate various data augmentation techniques and show the relevance of different augmentations, but also an order dependence of the augmentations. Grill~\etalcite{grill2020neurips} build upon earlier work \cite{chen2020icml} and proved that negative examples are not strictly necessary for pre-training. Zbontar~\etalcite{zbontar2021icml} extends this idea of using only positive examples by measuring the cross-correlation between two augmented views of the same image.

Recently, He~\etalcite{he2019iccv} challenged the need for pre-training and showed that, given enough iteration and data, a randomly initialized network can reach the same performance. Their work also confirms that pre-training is an effective way to reduce the need for labeled data and time to converge.
McCormac~\etalcite{mccormac2017iccv} compare the model pre-trained on ImageNet versus the model pre-trained on their synthetic RGB-D dataset, whose domain is aligned with the target dataset and task. Their domain-specific pre-training performs better than the ImageNet pre-training, especially when using depth information. 

In contrast to the supervised approaches, we aim at exploiting the large amount of unlabeled images that we can record using robotic systems. For the self-supervised pre-training, we use Barlow Twins~\cite{zbontar2021icml}. We evaluate the importance of pre-training directly in the agricultural domain, instead of adopting a general pre-training on ImageNet. In our work, we show the advantages of domain-specific augmentations, for which there is not much literature~\cite{cubuk2019cvpr, ratner2017neurips} and we examine how their order and application may influence the performance of the final system on the downstream task. 
%Our work proceeds towards a similar path, evaluating if domain-specific pre-training on nonsynthetic data can outperform the ImageNet pre-training. We also examine the advantages of this approach, data and computation wise, and the importance of domain-specific augmentations. 

%%%%%%%%%%%%%%%%%%%%%%%%%%%%%%%%%%%%%%%%%%%%%%%%%%%%%%%%%%%%%%%%%%%%%%%%%%%%%%%%
\section{Our Approach}
\label{sec:main}

%% Describe your approach. It is okay to divide the main section
%%  into a few subsections (e.g., 2-4 subsections)
We aim at learning an abstract representation that will serve as a starting point for further learning tasks in the domain of the perception of plants. By deploying robots in the fields, we can quite easily collect a large amount of \emph{unlabeled} data. This offers the potential to build systems to train networks in a self-supervised fashion. For our perception task, we pre-train the network encoder following Barlow Twins (BT) proposed by Zbontar~\etalcite{zbontar2021icml}. We decided for BT to avoid the problem of negative pairs since in the agricultural domain most of the images represent plants; the pre-training approach, as well as the architecture for the tasks, is not our main focus. We propose a domain-specific augmentation policy to boost the performance of the final system. We evaluate our pre-training on semantic and leaf instance segmentation.

\subsection{Barlow Twins}
\label{sec:arch}

BT learns representations in a self-supervised fashion via redundancy reduction. Here. we briefly summarize its relevant parts and refer for more details to the original paper~\cite{zbontar2021icml}. It uses a siamese network with shared weights, which can be seen in~\figref{fig:arch}. The two inputs are two different augmentations of the same input image. The encoder is a ResNet50~\cite{he2016cvpr} without the final classification layer, followed by a projector network. The projector network has two identical blocks --- linear layer, batch normalization, and rectified linear units --- followed by one linear layer.  

Zbontar~\etal build for each input image $I$ two augmented views $I_1$, $I_2$ that are fed into the network to produce two distinct embeddings $z_1$, $z_2\in\mathbb{R}^{\mathcal{D}}$. They compute the loss directly on $z_1$ and $z_2$. The first step is to construct the cross-correlation squared matrix $\mathcal{C} \in \mathbb{R}^{\mathcal{D} \times \mathcal{D}}$ from the embeddings normalized over the batch dimension. This matrix has values between $-1$ (anti-correlation) and $1$ (correlation). Then, the loss is:
\begin{equation}
\label{eq:btloss}
\mathcal{L_{BT}} \triangleq \sum_i (1 - \mathcal{C}_{ii})^2 + \lambda \sum_i \sum_{j \neq i} \mathcal{C}_{ij}^2~\text{,}
\end{equation}

\noindent where $\lambda$ is a weight to trade-off two parts: the invariance term, which forces the diagonal elements of $\mathcal{C}$ to be 1, and the redundancy reduction term, which forces all the non-diagonal elements of $\mathcal{C}$ to be 0. 

\subsection{Augmentations} 
\label{sec:aug}
Augmentations play a fundamental role in self-supervised learning. The stronger they are, the more the network focuses on relevant and stable features to represent the images. We use our augmentation policy as common in the literature~\cite{grill2020neurips} plus domain-specific knowledge. In \figref{fig:augmentations}, we show the result of each augmentation applied to one sample image for illustration purposes. Our augmentations are:

\subsubsection{Affine Transformation}
\label{sec:aug_affine}
The affine transformation, \mbox{$\mathbf{T}_\text{affine} \in \mathbb{R}^{3 \times 3}$} rotates, translates, scales, and shears the input image. It makes the network invariant to such transformations which are common when working with robots of different sizes and cameras. More specifically, $\mathbf{T}_\text{affine}$ is given by 

\begin{equation}
	\mathbf{T}_{\text{affine}} = \begin{bmatrix}
	\mathbf{A} & \b{t} \\ \b{0}^T & 1 
\end{bmatrix}	 ~\text{,}
\end{equation}

\noindent where \mbox{$\mathbf{A} \in \mathbb{R}^{2 \times 2}$} contains an isotropic scaling factor in $[0.5,2]$, a rotation in $[-\pi, \pi]$ and the shearing along the two axes randomly sampled in $[0.25, 0.75]$, while \mbox{$\bt \in \mathbb{R}^2$} is a translation vector with each component \mbox{$t_x, t_y \in [-0.25 \cdot W, 0.25 \cdot H]$}, where $W$ and $H$ are the image width and height respectively.

\subsubsection{Color Jittering}
\label{sec:aug_color}

Color jittering changes the brightness, contrast, hue, and saturation of the image. Instead of the symmetrical range of values for the hue $(-0.1, 0.1)$ from the literature, we use $(0, 0.125)$ as range. This transformation is crucial for the classification of ill or damaged plants, where color is a dominant discriminator~\cite{tuba2017wseas}. 

\subsubsection{Gaussian Blur}
\label{sec:aug_gauss}

We blur the image using a random standard deviation $\in [0.1, 2]$. The purpose of this augmentation is to help the network focus on the image structure across different scales and resolutions.

\subsubsection{Mixing}
\label{sec:aug_mix}

Zhang~\etalcite{zhang2018iclr} propose to mix two images via linear interpolation, we instead use a single image $I$. We create two copies of the image $I$, one flipped on the x-axis $I_x$ and one on the y-axis $I_y$. We sample each pixel in the augmented image from $I$, $I_x$, or $I_y$ using a uniform probability over the three images. This can simulate motion due to the wind or water uptake, or holes eaten by insects.

\subsubsection{Random Erasing}
\label{sec:aug_re}
Random erasing~\cite{zhong2020aaai} selects multiple rectangles inside of the image and substitutes the pixels' values with random values in $[0,255]$. Given the minimum percentage of the image that has to be removed, it picks rectangles of different sizes and aspect ratios until the deleted area is at least the minimum desired area. We slightly change the implementation to enforce the use of multiple rectangles with respect to a big one. We use this augmentation to make the network less sensitive to occlusions and shadows.

\subsubsection{Background Invariance}
\label{sec:aug_bg}

This transformation cuts plants from the current image and pastes them into a different soil background. Specifically, we perform the following steps:

\begin{enumerate}[label=(\roman*)]
	\item Compute a normalized image as \begin{equation}
		I_{\textit{norm}}(u,v) = \frac{I(u,v) - \mu_I}{\sigma_I + \epsilon}~\text{,}
	\end{equation}where $u,v$ are pixel coordinates, \mbox{$\mu_I \in \mathbb{R}^3$} and \mbox{$\sigma_I \in \mathbb{R}^3$} are mean and standard deviation of $I$, and $\epsilon = 10^{-8}$.
	\item Compute the vegetation mask $M$ following Woebbecke \etalcite{woebbecke1994asabe}. Specifically, $M$ is given by
		\begin{equation}
		M = 2 I_G - I_R - I_B~\text{,}
	\end{equation}where $I_R$,$I_G$ and $I_B$ are the color channels of $I_{\textit{norm}}$.
	\item Convert $M$ to a binary mask using a threshold $\theta$, \ie, all pixels above $\theta$ are set to 1, the others are set to 0. 	
	\item Refine $M$ using $2$ rounds of erosion with kernel size $(2,2)$, $4$ rounds of dilation with kernel size $(6,6)$. 
	\item Cut the vegetation and paste it at a random location on a random soil image from a dataset of images whose vegetation mask $M$ is below a given threshold, \ie, less than 5\% of the image. 
\end{enumerate}
All the augmentations are applied with a certain probability, tuned from the results in \secref{subsec:eval_trans}. 
\begin{figure}[t]
	\centering
	\includegraphics[scale=0.5]{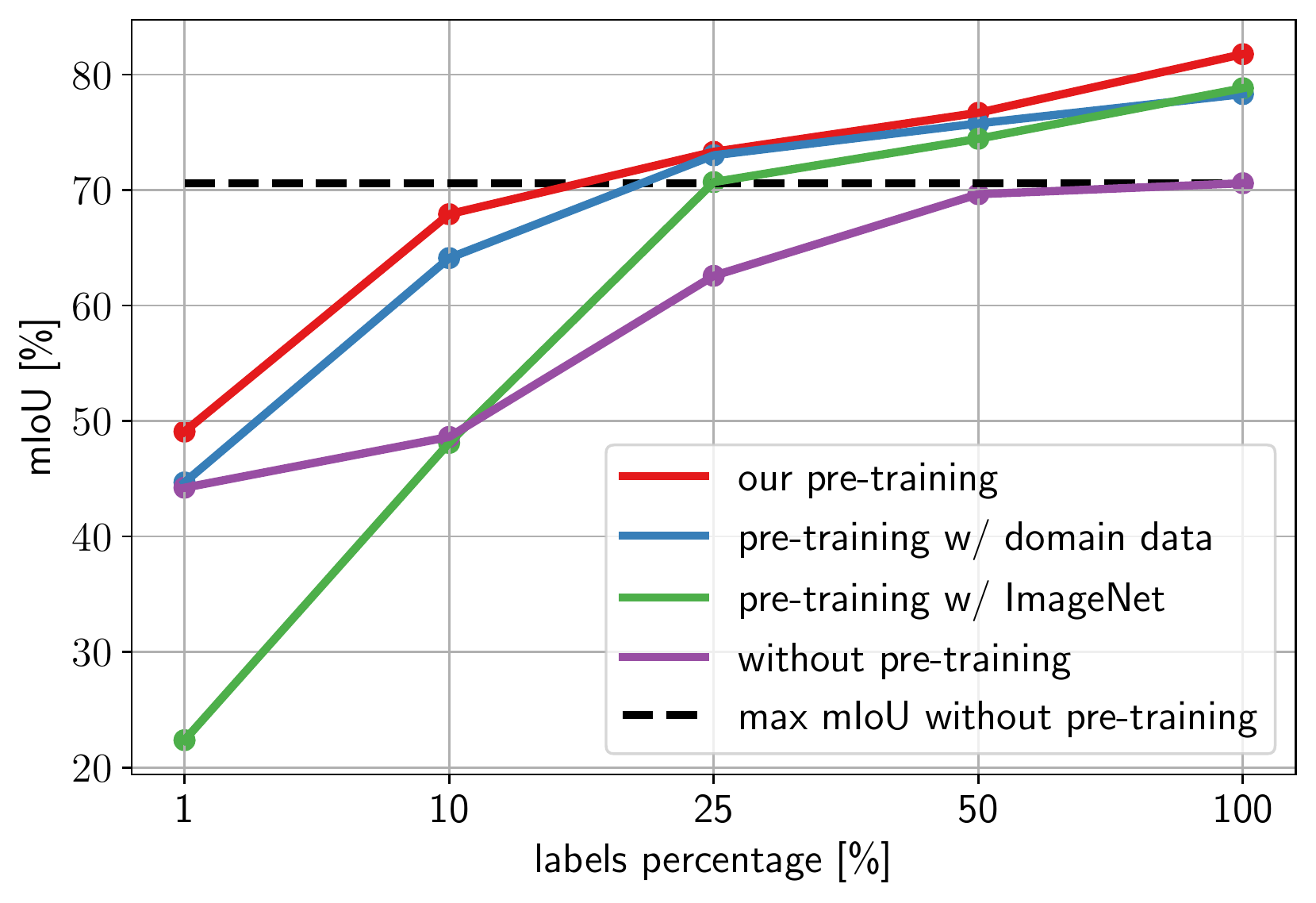}
	\caption{\label{fig:me_vs_imagenet} Comparison of the mIoU with different amounts of labels after fine-tuning for 100 epochs (semantic segmentation). The number of images for each label percentage is: 14 for 1\%, 140 for 10\%, 362 for 25\%, 724 for 50\%, and 1450 for 100\%.}
	\vspace{-0.2in}
\end{figure}

\section{Downstream Tasks}
\label{sec:eval}
In an application, the pre-trained models are fine-tuned on specific downstream tasks. Since the pre-training approach is backbone invariant, we pre-trained ResNet50 for the semantic segmentation and an ERFNet~\cite{romera2018tits} encoder for the leaf instance segmentation. For both tasks, we use ERFNet-like decoders.

\begin{table}[t]
\vspace{0.05in}
\caption{Comparison of average precision (AP) and recall (AR) on plants (p) and leaves (l) for the three main approaches. The number of images for label percentage is: 7 for 1\%, 74 for 10\%, 186 for 25\%, 373 for 50\%, and 746 for 100\%.}
\label{tab:lis_compare}
\centering
\begin{tabular}{c c c c c}
\toprule

\multicolumn{1}{c}{Pre-Training} & \multicolumn{1}{c}{\textbf{$AP_{p}$ [\%]}} & \multicolumn{1}{c}{\textbf{$AR_{p}$ [\%]}} & \multicolumn{1}{c}{\textbf{$AP_{l}$ [\%]}} & \multicolumn{1}{c}{\textbf{$AR_{l}$ [\%]}} \\ \midrule

\multicolumn{5}{c}{100\% of labels} \\ \midrule

\multicolumn{1}{c}{none} & \multicolumn{1}{c}{ 54.3 } & \multicolumn{1}{c}{ 60.5 } & \multicolumn{1}{c}{48.7} & \multicolumn{1}{c}{68.3} \\

\multicolumn{1}{c}{ImageNet} & \multicolumn{1}{c}{ 55.1 } & \multicolumn{1}{c}{ 61.2 } & \multicolumn{1}{c}{59.7} & \multicolumn{1}{c}{68.9} \\

\multicolumn{1}{c}{ours} & \multicolumn{1}{c}{ \textbf{55.6} } & \multicolumn{1}{c}{ \textbf{62.9} } & \multicolumn{1}{c}{\textbf{64.4}} & \multicolumn{1}{c}{\textbf{69.2}} \\ \midrule

\multicolumn{5}{c}{50\% of labels} \\ \midrule

\multicolumn{1}{c}{none} & \multicolumn{1}{c}{ 50.3 } & \multicolumn{1}{c}{ 59.0 } & \multicolumn{1}{c}{45.6} & \multicolumn{1}{c}{60.1} \\

\multicolumn{1}{c}{ImageNet} & \multicolumn{1}{c}{ 52.4 } & \multicolumn{1}{c}{ 60.1 } & \multicolumn{1}{c}{52.7} & \multicolumn{1}{c}{61.5} \\

\multicolumn{1}{c}{ours} & \multicolumn{1}{c}{ \textbf{54.6} } & \multicolumn{1}{c}{ \textbf{60.8} } & \multicolumn{1}{c}{\textbf{54.6}} & \multicolumn{1}{c}{\textbf{62.7}} \\ \midrule

\multicolumn{5}{c}{25\% of labels} \\ \midrule

\multicolumn{1}{c}{none} & \multicolumn{1}{c}{ 48.0 } & \multicolumn{1}{c}{ 55.2 } & \multicolumn{1}{c}{42.0} & \multicolumn{1}{c}{46.1} \\

\multicolumn{1}{c}{ImageNet} & \multicolumn{1}{c}{ 50.2 } & \multicolumn{1}{c}{ 56.1 } & \multicolumn{1}{c}{50.6} & \multicolumn{1}{c}{56.6} \\

\multicolumn{1}{c}{ours} & \multicolumn{1}{c}{ \textbf{50.9} } & \multicolumn{1}{c}{ \textbf{56.9} } & \multicolumn{1}{c}{\textbf{53.8}} & \multicolumn{1}{c}{\textbf{60.6}} \\ \midrule

\multicolumn{5}{c}{10\% of labels} \\ \midrule

\multicolumn{1}{c}{none} & \multicolumn{1}{c}{ 46.6 } & \multicolumn{1}{c}{ 54.0 } & \multicolumn{1}{c}{20.7} & \multicolumn{1}{c}{39.6} \\

\multicolumn{1}{c}{ImageNet} & \multicolumn{1}{c}{ 46.8 } & \multicolumn{1}{c}{ 53.7 } & \multicolumn{1}{c}{29.5} & \multicolumn{1}{c}{38.4} \\

\multicolumn{1}{c}{ours} & \multicolumn{1}{c}{ \textbf{48.0} } & \multicolumn{1}{c}{ \textbf{54.2} } & \multicolumn{1}{c}{\textbf{42.5}} & \multicolumn{1}{c}{\textbf{49.2}} \\ \midrule

\multicolumn{5}{c}{1\% of labels} \\ \midrule

\multicolumn{1}{c}{none} & \multicolumn{1}{c}{ 0.0} & \multicolumn{1}{c}{0.0 } & \multicolumn{1}{c}{0.1} & \multicolumn{1}{c}{0.3} \\

\multicolumn{1}{c}{ImageNet} & \multicolumn{1}{c}{ 0.0 } & \multicolumn{1}{c}{ 0.0 } & \multicolumn{1}{c}{0.4} & \multicolumn{1}{c}{0.2} \\

\multicolumn{1}{c}{ours} & \multicolumn{1}{c}{ \textbf{1.1} } & \multicolumn{1}{c}{ \textbf{8.3} } & \multicolumn{1}{c}{\textbf{0.9}} & \multicolumn{1}{c}{\textbf{5.6}} \\
\bottomrule
\end{tabular}
\vspace{-0.2in}
\end{table}

\subsection{Semantic Segmentation}
\label{sec:sem_seg}
Semantic segmentation predicts a class for each pixel of the image, in our application example, crop, weed, and background. The decoder outputs an image $H \times W \times C_{out}$, where $C_{out}$ is the number of semantic classes, $H$ and $W$ are the height and width of the input image. Instead of connecting the decoder at the end of the ResNet50, we discard the last two layers to preserve more spatial information. We can initialize the remaining part with the pre-trained weights without changing anything. We follow Rahman~\etalcite{rahman2016icsv} to directly optimize the IoU.

\subsection{Leaf Instance Segmentation} 
Leaf instance segmentation predicts a pixel-wise mask for each leaf. Such task allows discovering not only the shape and size of individual leaves but also counting them, which is fundamental in determining the growth stage of the plant~\cite{farjon2021fps}. 
We use the network and loss proposed by Weyler~\etalcite{weyler2022wacv}. One decoder predicts the center locations of each leaf, the other the offsets pointing at the specific leaf and plant center plus clustering parameters for the post-processing. The predicted and the ground truth masks are fed to the Lov\'{a}sz Hinge Loss~\cite{berman2018cvpr}. For more details, refer to the original paper~\cite{weyler2022wacv}. 

%%%%%%%%%%%%%%%%%%%%%%%%%%%%%%%%%%%%%%%%%%%%%%%%%%%%%%%%%%%%%%%%%%%%%%%%%%%%%%%%
\section{Experimental Evaluation}
\label{sec:exp}

%% Repeat the main focus/objective with one single(!) sentence starting with:
%
%The main focus of this work is a  \dots
The focus of this work is showing that a domain-specific self-supervised pre-training together with augmentation policy has the potential to perform better than the commonly used supervised pre-training on ImageNet. We show this for images from the agricultural robotics domain.
We present our experiments to show the capabilities of our pre-training and to support our key claims, which are:
(i)~pre-training on plant images improves the performance of the downstream tasks,
(ii)~using domain-specific pre-training can further reduce the number of labels needed, and
(iii)~the augmentation policy needs to be domain-specific and consider the design of each augmentation to apply them in the best order.

\subsection{Experimental Setup}

Our best model has the encoder pre-trained for 250 epochs, with batch size 128, learning rate $2\cdot{10}^{-4}$, and a weight decay of ${10}^{-6}$. We use 18,000 images from four different locations (Ancona in Italy, Bonn and Stuttgart in Germany, and Eschlikon in Switzerland, see also the dataset paper~\cite{chebrolu2017ijrr}) for pre-training. We compare it with a publicly available supervised pre-training on ImageNet.
For the semantic segmentation task, we use a dataset that contains 2,148 images: 1,450 for training, 478 for validation, and 220 for testing. For the leaf instance segmentation, we use a dataset with 1,316 images; 746 for training, 292 for validation, and 278 for testing. We pre-trained on a single NVIDIA RTX A6000 GPU and fine-tuned on a single Quadro RTX 5000 GPU. We plan to publish the datasets used both for pre-training and fine-tuning.

We evaluate the results on semantic segmentation using the mean intersection over union (mIoU)~\cite{everingham2010ijcv} over soil, crop, and weed. For the leaf instance segmentation task, we report the average precision (AP) and recall (AR), and the absolute difference in count ($|$DiC$|$) of the leaves. 
\begin{figure}[t]
\vspace{0.1in}
	\centering
	\includegraphics[scale=0.5]{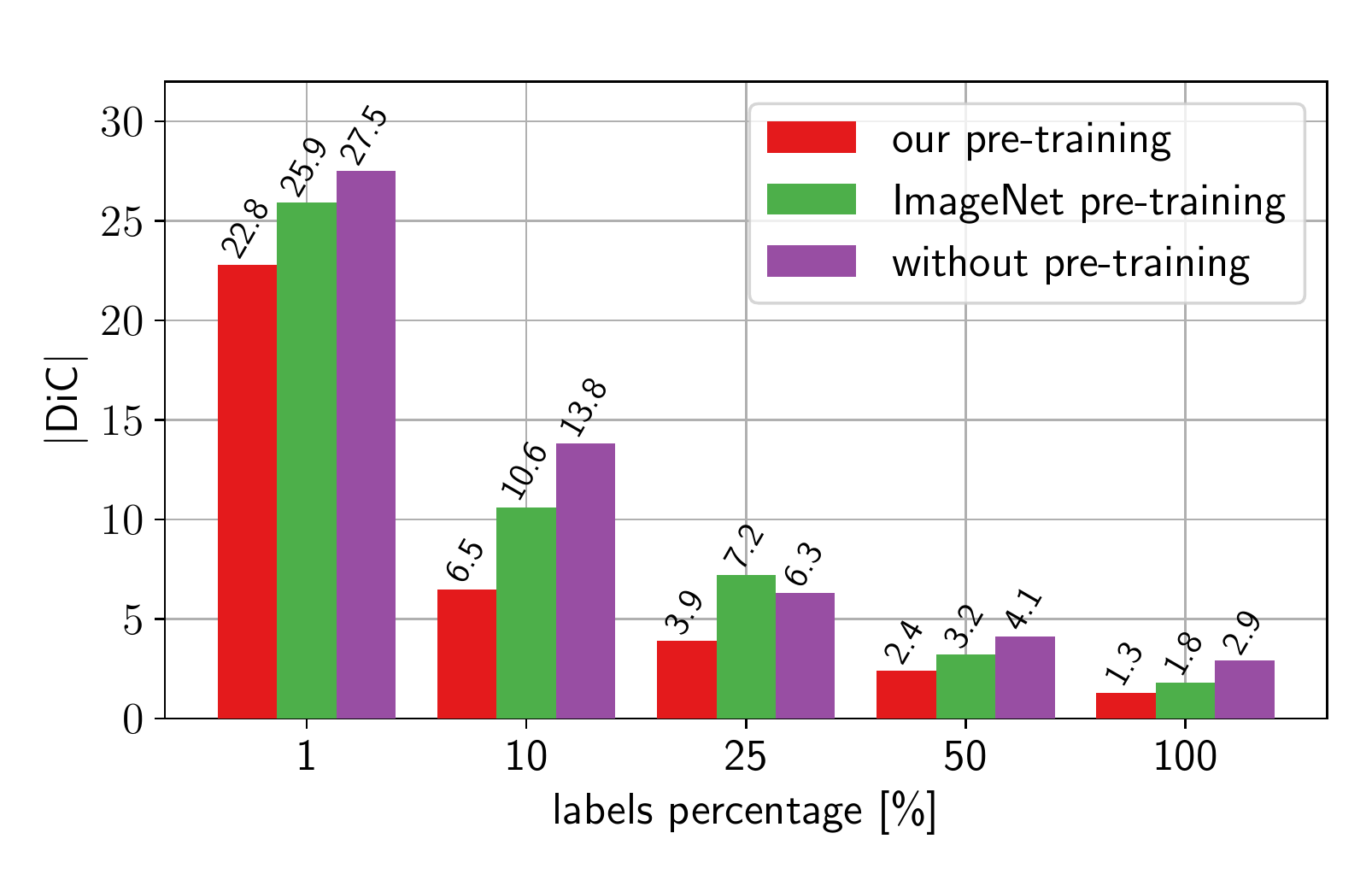}
	\caption{\label{fig:DiC}The average $\mid$DiC$\mid$ for the three approaches with increasing number of labeled images. The lower the better.}
\vspace{-0.2in}
\end{figure}
\subsection{Our Pre-training vs. Non-specific Pre-training}

The first experiment analyzes how self-supervised domain-specific pre-training improves the effectiveness and decreases the labeled images, time, and computational resources needed. We fine-tuned our model and the model pre-trained on ImageNet with different amounts of labels.

\textbf{Semantic Segmentation.} \figref{fig:me_vs_imagenet} suggests that when using a sufficient number of labels, different pre-training strategies perform similarly. The fewer labels we use, the wider the gap. The ImageNet pre-training requires more labeled data to adjust to the agriculture domain. Our pre-training performs better %- from $+27\%$ using $14$ images to $+3\%$ using $1450$ images - 
requiring less data for pre-training i.e. 18,000 images against the 1,281,167 from ImageNet, and epochs i.e. 250 against 1,000. Only for this experiment, we also pre-trained on domain-specific data with the augmentations from the literature. The results confirm that domain-specific augmentations are a key component to obtain the best performance. 

\textbf{Leaf Instance Segmentation.} \tabref{tab:lis_compare} confirms the utility of pre-training on domain-specific data to boost the performance and reduce the number of labeled images needed. Our pre-training boosts every metric in every scenario. In \figref{fig:DiC}, the difference in count shows a similar trend. When using less than 10 images none of the approaches can properly segment the leaves, but using 74 images (10\%) our pre-training can already reduce the uncounted leaves to $\approx\,$6 per image (each image can have between 2 and 5 plants in it).

\begin{figure}[t]
\centering
\includegraphics[scale=0.475]{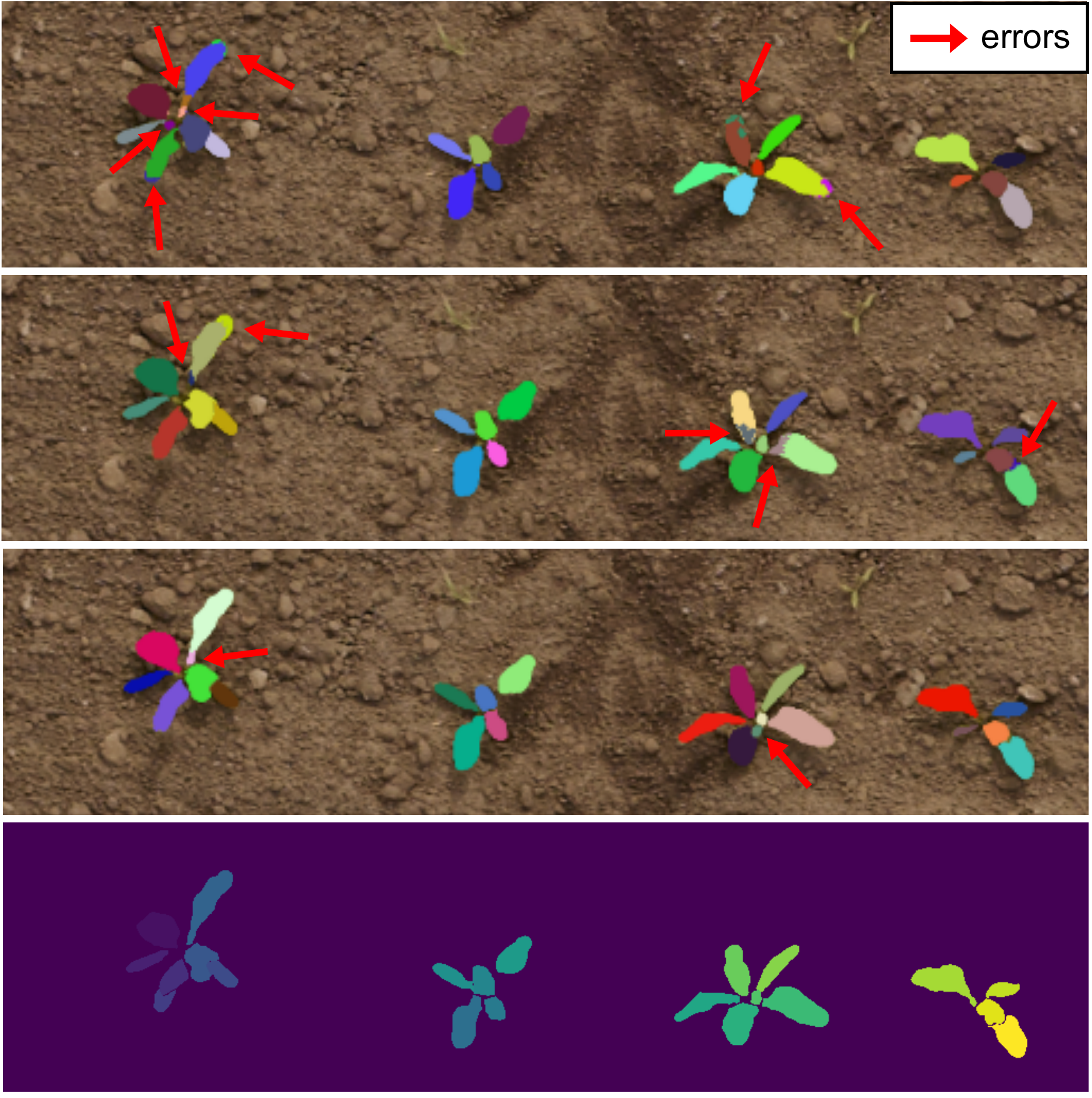}
\caption{\label{fig:leaf_instance} Leaf instance segmentation with 50\% of the labels. From top to bottom, we show the result with no pre-training, ImageNet and our pre-training, then the ground truth.}
\vspace{-0.2in}
\end{figure}
%The model pre-trained on ImageNet has trained for 1000 epochs with batch size 1024.
\begin{figure}[t]
	\centering
	\includegraphics[scale=0.5]{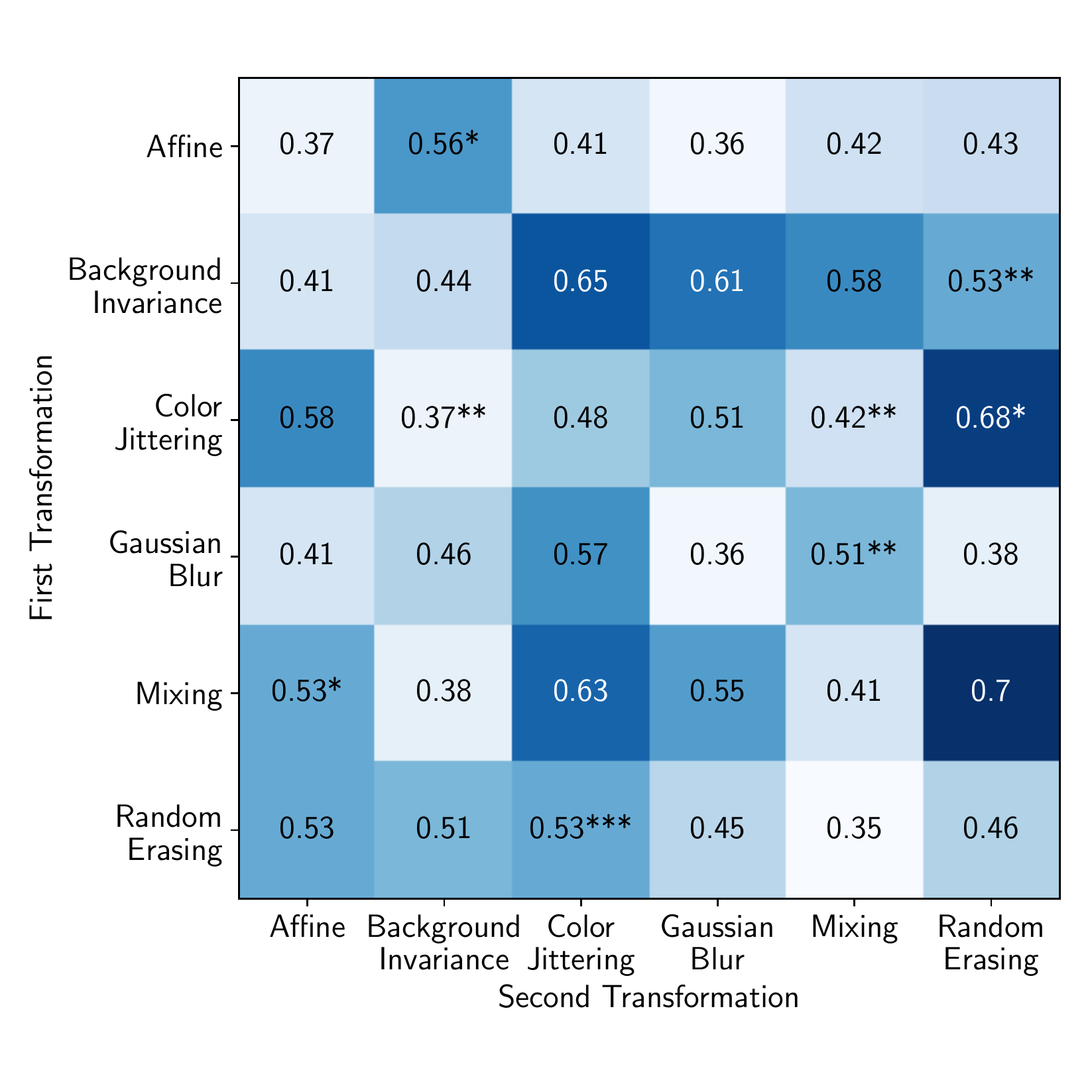}
	\vspace{-4mm}
	\caption{\label{fig:transformations_table}The mIoU for different combinations of transformations. We fine-tuned with minimum 100 epochs or until convergence: (*) 10 extra epochs; (**) 40 extra epochs; (***) 100 extra epochs.}
\end{figure}

%\subsection{How much does the pre-training help?}
\subsection{Our Pre-training vs. No Pre-training}

%The literature has already answered to this question multiple times, 
This experiment aims to compare the results and computational resources when using the pre-trained model with respect to training the network after a random initialization.

\textbf{Semantic Segmentation.} In \figref{fig:me_vs_imagenet}, we see how pre-training boosts the performance when using the same routine and number of labeled images. Our pre-trained model performs better using up to 25\% of the annotated data. With less than 10\% of the labeled data, pre-training on ImageNet hurts the performance. The reason may be that the network expects to see objects from the ImageNet dataset distribution and requires the labeled data to adapt to the agricultural domain. Our pre-training does not suffer from this issue.

\textbf{Leaf Instance Segmentation.} Training the randomly initialized network with few labeled images deteriorates rapidly the performance. \figref{fig:DiC} and \tabref{tab:lis_compare} show that we can obtain the same performance using half of the data. Our pre-training boosts all of the metrics, and it produces instance masks using only 7 images (1\%). 

%\subsection{Which transformations are better?}
\subsection{Relevance and Order of the Augmentations}
\label{subsec:eval_trans}

We used a shorter training routine to analyze which augmentations work better for the plant domain. Each combination has been pre-trained for 50 epochs on a subset of the pre-training dataset, then fine-tuned on a subset of the dataset for semantic segmentation. Chen \etalcite{chen2020icml} did a similar experiment whose results agree with ours.

In \figref{fig:transformations_table}, we see that changing the transformation's order impacts the mIoU and the training time. The combinations that took more time are those that make the task much harder, i.e., if we first apply color jittering and then background invariance, it will be challenging to correctly identify the plants, with the vegetation mask corrupted by the changes in color. Focusing on the highest-performing combinations we see that swapping them leads to lower values, confirming that the order in which they are applied is a key aspect when designing the augmentation policy. On the diagonal, where we apply only one augmentation, the mIoU values are all in the mid-lower range, the highest values being the strongest augmentations. This pattern is also visible in the other combinations; strong augmentations such as random erasing and mixing lead to better performance, not always at the price of longer training time.

\begin{table}[t]
\caption{The mIoU [\%] after fine-tuning (100 epochs on semantic segmentation) with 20, 40, 60, 80 and 100 epochs of pre-training using only the color transformation.}
\label{tab:colors} 
\centering
\begin{tabular}{cccccc}
\toprule
\multicolumn{1}{c}{} & \multicolumn{5}{c}{\textbf{pre-training epochs}} \\
\cline{2-6}
\multicolumn{1}{c}{\textbf{Approach}} & \multicolumn{1}{c}{\textbf{20}} & \multicolumn{1}{c}{\textbf{40}} & \multicolumn{1}{c}{\textbf{60}} & \multicolumn{1}{c}{\textbf{80}} & \multicolumn{1}{c}{\textbf{100}}  \rule{0pt}{3ex}  \\
\midrule
\multicolumn{1}{c}{standard} & \multicolumn{1}{c}{\textbf{23.44}} & \multicolumn{1}{c}{\textbf{23.77}} & \multicolumn{1}{c}{23.89} & \multicolumn{1}{c}{19.92} & \multicolumn{1}{c}{13.89} \\ 
\multicolumn{1}{c}{our}  & \multicolumn{1}{c}{17.61} & \multicolumn{1}{c}{21.11} & \multicolumn{1}{c}{\textbf{31.28}} & \multicolumn{1}{c}{\textbf{32.52}}  & \multicolumn{1}{c}{\textbf{42.80}}\\  \bottomrule
\end{tabular}
\vspace{-0.1in}
\end{table}

As explained in \secref{sec:aug_color}, we changed the usual parameters for the color jittering augmentation. To evaluate if this choice makes a difference we pre-trained our encoder only with color jittering and used the weights at different stages for the semantic segmentation task. We demonstrate in \tabref{tab:colors} that a longer pre-training period with the standard color augmentation degrades performance. One reason could be that the augmentation is so strong that the network does not take color into account anymore, making the semantic segmentation task harder. Our augmentation instead provides better performance the more the encoder is pre-trained.

%\subsection{How many epochs?}
\subsection{Influence of Pre-training Length}

\begin{figure}[t]
	\centering
	\includegraphics[scale=0.5]{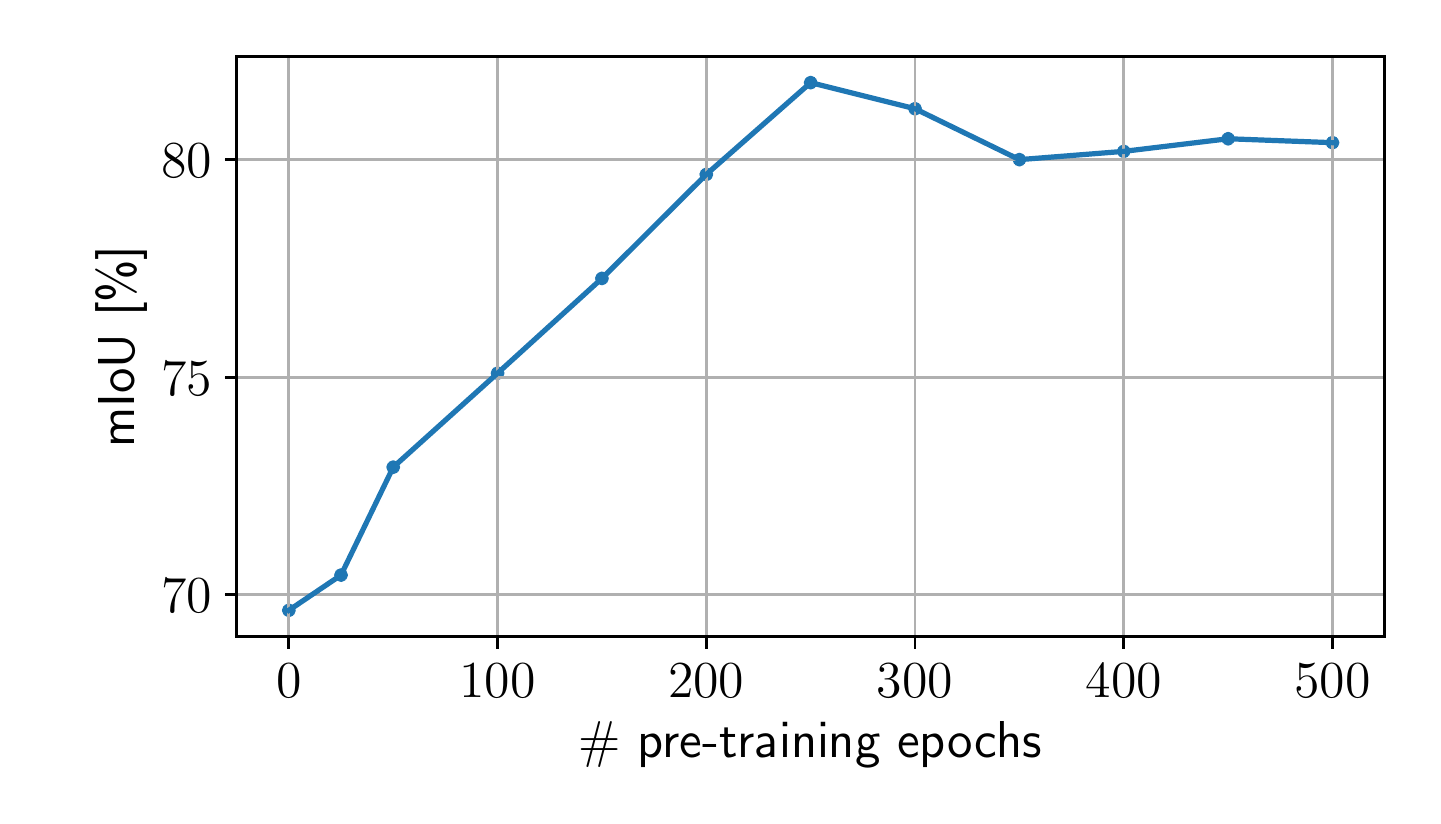}
	\caption{\label{fig:miou_on_epochs}The mIoU after fine-tuning for 100 epochs using different pre-training epochs. There is no improvement after 250 epochs. The starting value is obtained with a randomly initialized network.}
\end{figure}

We pre-trained up to 500 epochs, evaluating intermediate checkpoints on the semantic segmentation task. We want to determine how many epochs are needed to achieve satisfying results and to find out how much we can improve by continuing training. This allows us to find a compromise between performance and computational resources. In \figref{fig:miou_on_epochs} we show that the mIoU increases until 250 epochs, where we see a diminishing effect. Therefore, if not otherwise specified, we pre-train for 250 epochs in our experiments.

% plottone di IoU con vari checkpoints 

%\subsection{Why not using always every transformation?}
\subsection{Ablation on Augmentations' Probabilities}

%i am just gonna assume the weighted pre training is better, let's hope for it fuck 
%\begin{table}[]
%\centering
%\begin{tabular}{l c}
%\multicolumn{1}{l|}{Augmentation} & \multicolumn{1}{c}{Probability} \\ \hline \hline
%\multicolumn{1}{l|}{Affine} & \multicolumn{1}{c}{0.8}  \\ \hline
%\multicolumn{1}{l|}{Background Invariance}  & \multicolumn{1}{c}{0.8}  \\ \hline 
%\multicolumn{1}{l|}{Color Jittering}  & \multicolumn{1}{c}{1.0}  \\ \hline 
%\multicolumn{1}{l|}{Gaussian Blur}  & \multicolumn{1}{c}{0.9}  \\ \hline 
%\multicolumn{1}{l|}{Mixing}  & \multicolumn{1}{c}{0.9}  \\ \hline 
%\multicolumn{1}{l|}{Random Erasing}  & \multicolumn{1}{c}{1.0}  \\ \hline 
%\end{tabular}
%\caption{\label{tab:prob_aug} The probabilities we assign to each augmentation.}
%\end{table}

Each augmentation is applied with a probability based on the results of the previous experiments to increase variance. For evaluation of the effectiveness of our policy, we compare it against a pre-training in which all augmentations are always applied.

\begin{table}[t]
\caption{Comparison of mIoU, mean precision (mP), and mean recall (mR) after fine-tuning for 100 epochs (semantic segmentation) with our augmentation's probabilities vs. applying always all the augmentations.}
\label{tab:me_vs_flat_prob}
\centering
\begin{tabular}{c c c c}
\toprule
\multicolumn{1}{c}{Approach} & \multicolumn{1}{c}{\textbf{mIoU [\%]}} & \multicolumn{1}{c}{\textbf{mP [\%]}} & \multicolumn{1}{c}{\textbf{mR [\%]}} \\ \midrule
\multicolumn{1}{c}{all augmentations} & \multicolumn{1}{c}{78.09} & \multicolumn{1}{c}{88.61} & \multicolumn{1}{c}{87.68} \\
\multicolumn{1}{c}{our policy}  & \multicolumn{1}{c}{\textbf{81.77}} & \multicolumn{1}{c}{\textbf{91.07}}  & \multicolumn{1}{c}{\textbf{89.99}}  \\ \bottomrule
\end{tabular}
\vspace{-0.1in}
\end{table}

The results in \tabref{tab:me_vs_flat_prob} show that the probabilities we assign to each augmentation result in higher performance on all metric evaluations. We propose to assign a probability of 1.0 to color jittering and random erasing, 0.9 to gaussian blur and mixing, and 0.8 to background invariance and affine transform.

\vspace{-0.1cm}
%%%%%%%%%%%%%%%%%%%%%%%%%%%%%%%%%%%%%%%%%%%%%%%%%%%%%%%%%%%%%%%%%%%%%%%%%%%%%%%%
\section{Conclusion}
\label{sec:conclusion}

In this paper, we presented an approach to exploit a vast quantity of unlabeled images from the agricultural domain to learn useful representations in a self-supervised fashion. Our experiments rely on domain-specific data and domain-specific augmentations during the pre-training. This allows us to successfully use our pre-training for different downstream tasks obtaining good performance using less labeled images. We implemented and evaluated our pre-trainings on two tasks, semantic and leaf instance segmentation in the agriculture domain. We compared our results with those obtained without pre-training and pre-training on ImageNet and supported all claims made in this paper. The experiments suggest that pre-training on a domain-specific dataset and exploiting domain knowledge to define the augmentation policy can reduce the number of labeled data required to achieve the same performances as without pre-training. 

\bibliographystyle{plain_abbrv}

\bibliography{glorified,new}

\begin{thebibliography}{10}

\bibitem{ayalew2020eccv}
T.~Ayalew, J.~Ubbens, and I.~Stavness.
\newblock Unsupervised domain adaptation for plant organ counting.
\newblock In {\em Proc.~of the Europ.~Conf.~on Computer Vision (ECCV)}, 2020.

\bibitem{berman2018cvpr}
M.~Berman, A.R. Triki, and M.B. Blaschko.
\newblock The lov{\'a}sz-softmax loss: A tractable surrogate for the
  optimization of the intersection-over-union measure in neural networks.
\newblock In {\em Proc.~of the IEEE/CVF Conf.~on Computer Vision and Pattern
  Recognition (CVPR)}, 2018.

\bibitem{buzzy2020sensors}
M.~Buzzy, V.~Thesma, M.~Davoodi, and J.~Mohammadpour~Velni.
\newblock Real-time plant leaf counting using deep object detection networks.
\newblock {\em Sensors}, 20(23):6896, 2020.

\bibitem{caron2021neurips}
M.~Caron, I.~Misra, J.~Mairal, P.~Goyal, P.~Bojanowski, and A.~Joulin.
\newblock Unsupervised learning of visual features by contrasting cluster
  assignments.
\newblock In {\em Proc.~of the Conference on Neural Information Processing
  Systems (NeurIPS)}, 2021.

\bibitem{chebrolu2017ijrr}
N.~Chebrolu, P.~Lottes, A.~Schaefer, W.~Winterhalter, W.~Burgard, and
  C.~Stachniss.
\newblock {Agricultural Robot Dataset for Plant Classification, Localization
  and Mapping on Sugar Beet Fields}.
\newblock {\em Intl.~Journal~of Robotics Research (IJRR)}, 36:1045--1052, 2017.

\bibitem{chen2020icml}
T.~Chen, S.~Kornblith, M.~Norouzi, and G.~Hinton.
\newblock {A Simple Framework for Contrastive Learning of Visual
  Representations}.
\newblock In {\em Proc.~of the Intl.~Conf.~on Machine Learning (ICML)}, 2020.

\bibitem{chen2020arxiv}
X.~Chen, H.~Fan, R.~Girshick, and K.~He.
\newblock Improved baselines with momentum contrastive learning.
\newblock {\em arXiv preprint}, 2003.04297, 2020.

\bibitem{cubuk2019cvpr}
E.D. Cubuk, B.~Zoph, D.~Mane, V.~Vasudevan, and Q.V. Le.
\newblock Autoaugment: Learning augmentation policies from data.
\newblock In {\em Proc.~of the IEEE/CVF Conf.~on Computer Vision and Pattern
  Recognition (CVPR)}, 2019.

\bibitem{deng2009cvpr}
J.~Deng, W.~Dong, R.~Socher, L.~Li, K.~Li, and L.~Fei-Fei.
\newblock Imagenet: A large-scale hierarchical image database.
\newblock In {\em Proc.~of the IEEE Conf.~on Computer Vision and Pattern
  Recognition (CVPR)}, 2009.

\bibitem{erhan2010jmlr}
D.~Erhan, Y.~Bengio, A.~Courville, P.A. Manzagol, P.~Vincent, and S.~Bengio.
\newblock Why does unsupervised pre-training help deep learning?
\newblock {\em Journal of Machine Learning Research}, 11(19):625--660, 2010.

\bibitem{everingham2010ijcv}
M.~Everingham, L.~Van~Gool, C.~Williams, J.~Winn, and A.~Zisserman.
\newblock {The Pascal Visual Object Classes (VOC) Challenge}.
\newblock {\em Intl.~Journal~of Computer Vision (IJCV)}, 88(2):303--338, 2010.

\bibitem{farjon2021fps}
G.~Farjon, Y.~Itzhaky, F.~Khoroshevsky, and A.~Bar-Hillel.
\newblock Leaf counting: Fusing network components for improved accuracy.
\newblock {\em Frontiers in plant science}, 12:575751, 2021.

\bibitem{fiorani2013arpb}
F.~Fiorani and U.~Schurr.
\newblock Future scenarios for plant phenotyping.
\newblock {\em Annual review of plant biology}, 64:267--291, 2013.

\bibitem{gogoll2020iros}
D.~Gogoll, P.~Lottes, J.~Weyler, N.~Petrinic, and C.~Stachniss.
\newblock {Unsupervised Domain Adaptation for Transferring Plant Classification
  Systems to New Field Environments, Crops, and Robots}.
\newblock In {\em Proc.~of the IEEE/RSJ Intl.~Conf.~on Intelligent Robots and
  Systems (IROS)}, 2020.

\bibitem{goerlich2021drones}
F.~G{\"o}rlich, E.~Marks, A.K. Mahlein, K.~K{\"o}nig, P.~Lottes, and
  C.~Stachniss.
\newblock Uav-based classification of cercospora leaf spot using rgb images.
\newblock {\em Drones}, 5(2):34, 2021.

\bibitem{grill2020neurips}
J.B. Grill, F.~Strub, F.~Altché, C.~Tallec, P.H. Richemond, E.~Buchatskaya,
  C.~Doersch, B.A. Pires, Z.D. Guo, M.G. Azar, B.~Piot, K.~Kavukcuoglu,
  R.~Munos, and M.~Valko.
\newblock Bootstrap your own latent: A new approach to self-supervised
  learning.
\newblock In {\em Proc.~of the Conference on Neural Information Processing
  Systems (NeurIPS)}, 2020.

\bibitem{halstead2021fps}
M.~Halstead, A.~Ahmadi, C.~Smitt, O.~Schmittmann, and C.~McCool.
\newblock Crop agnostic monitoring driven by deep learning.
\newblock {\em Frontiers in plant science}, 12:786702, 2021.

\bibitem{hartley2021plants}
Z.K. Hartley and A.P. French.
\newblock Domain adaptation of synthetic images for wheat head detection.
\newblock {\em Plants}, 10(12):2633, 2021.

\bibitem{he2020cvpr-mcfu}
K.~He, H.~Fan, Y.~Wu, S.~Xie, and R.~Girshick.
\newblock {Momentum Contrast for Unsupervised Visual Representation Learning}.
\newblock In {\em Proc.~of the IEEE/CVF Conf.~on Computer Vision and Pattern
  Recognition (CVPR)}, 2020.

\bibitem{he2019iccv}
K.~He, R.~Girshick, and P.~Dollar.
\newblock {Rethinking ImageNet Pre-training}.
\newblock In {\em Proc.~of the IEEE/CVF Intl.~Conf.~on Computer Vision (ICCV)},
  2019.

\bibitem{he2016cvpr}
K.~He, X.~Zhang, S.~Ren, and J.~Sun.
\newblock Deep residual learning for image recognition.
\newblock In {\em Proc.~of the IEEE/CVF Conf.~on Computer Vision and Pattern
  Recognition (CVPR)}, 2016.

\bibitem{liu2016nips}
M.Y. Liu and O.~Tuzel.
\newblock Coupled generative adversarial networks.
\newblock In {\em Proc.~of the Conference on Neural Information Processing
  Systems (NeurIPS)}, 2016.

\bibitem{long2015cvpr-fcnf}
J.~Long, E.~Shelhamer, and T.~Darrell.
\newblock {Fully Convolutional Networks for Semantic Segmentation}.
\newblock In {\em Proc.~of the IEEE Conf.~on Computer Vision and Pattern
  Recognition (CVPR)}, 2015.

\bibitem{lottes2018iros}
P.~Lottes, J.~Behley, N.~Chebrolu, A.~Milioto, and C.~Stachniss.
\newblock {Joint Stem Detection and Crop-Weed Classification for Plant-specific
  Treatment in Precision Farming}.
\newblock In {\em Proc.~of the IEEE/RSJ Intl.~Conf.~on Intelligent Robots and
  Systems (IROS)}, 2018.

\bibitem{lottes2018ral}
P.~Lottes, J.~Behley, A.~Milioto, and C.~Stachniss.
\newblock Fully convolutional networks with sequential information for robust
  crop and weed detection in precision farming.
\newblock {\em IEEE Robotics and Automation Letters (RA-L)}, 3:3097--3104,
  2018.

\bibitem{lottes2017iros}
P.~Lottes and C.~Stachniss.
\newblock Semi-supervised online visual crop and weed classification in
  precision farming exploiting plant arrangement.
\newblock In {\em Proc.~of the IEEE/RSJ Intl.~Conf.~on Intelligent Robots and
  Systems (IROS)}, 2017.

\bibitem{magistri2020iros}
F.~Magistri, N.~Chebrolu, and C.~Stachniss.
\newblock {Segmentation-Based 4D Registration of Plants Point Clouds for
  Phenotyping}.
\newblock In {\em Proc.~of the IEEE/RSJ Intl.~Conf.~on Intelligent Robots and
  Systems (IROS)}, 2020.

\bibitem{marks2022icra}
E.~Marks, F.~Magistri, and C.~Stachniss.
\newblock {Precise 3D Reconstruction of Plants from UAV Imagery Combining
  Bundle Adjustment and Template Matching}.
\newblock In {\em Proc.~of the IEEE Intl.~Conf.~on Robotics \& Automation
  (ICRA)}, 2022.

\bibitem{mccormac2017iccv}
J.~McCormac, A.~Handa, S.~Leutenegger, and A.J.Davison.
\newblock {SceneNet RGB-D: Can 5M Synthetic Images Beat Generic ImageNet
  Pre-training on Indoor Segmentation?}
\newblock In {\em Proc.~of the IEEE Intl.~Conf.~on Computer Vision (ICCV)},
  2017.

\bibitem{milioto2018icra}
A.~Milioto, P.~Lottes, and C.~Stachniss.
\newblock {Real-time Semantic Segmentation of Crop and Weed for Precision
  Agriculture Robots Leveraging Background Knowledge in CNNs}.
\newblock In {\em Proc.~of the IEEE Intl.~Conf.~on Robotics \& Automation
  (ICRA)}, 2018.

\bibitem{potena2016ias}
C.~Potena, D.~Nardi, and A.~Pretto.
\newblock Fast and accurate crop and weed identification with summarized train
  sets for precision agriculture.
\newblock In {\em Proc. of Intl.~Conf.~on Intelligent Autonomous Systems
  (IAS)}, 2016.

\bibitem{pretto2020ram}
A.~Pretto, S.~Aravecchia, W.~Burgard, N.~Chebrolu, C.~Dornhege, T.~Falck,
  F.~Fleckenstein, A.~Fontenla, M.~Imperoli, R.~Khanna, F.~Liebisch, P.~Lottes,
  A.~Milioto, D.~Nardi, S.~Nardi, J.~Pfeifer, M.~Popović, C.~Potena,
  C.~Pradalier, E.~Rothacker-Feder, I.~Sa, A.~Schaefer, R.~Siegwart,
  C.~Stachniss, A.~Walter, W.~Winterhalter, X.~Wu, and J.~Nieto.
\newblock {Building an Aerial-Ground Robotics System for Precision Farming}.
\newblock {\em IEEE Robotics and Automation Magazine (RAM)}, 28(3):29--49,
  2020.

\bibitem{rahman2016icsv}
M.A. Rahman and Y.~Wang.
\newblock {Optimizing Intersection-Over-Union in Deep Neural Networks for Image
  Segmentation}.
\newblock In {\em Proc.~of the Int. Symp. on Visual Computing}, 2016.

\bibitem{ratner2017neurips}
A.J. Ratner, H.R. Ehrenberg, Z.~Hussain, J.~Dunnmon, and C.~Ré.
\newblock Learning to compose domain-specific transformations for data
  augmentation.
\newblock In {\em Proc.~of the Conference on Neural Information Processing
  Systems (NeurIPS)}, 2017.

\bibitem{razavian2014arxiv}
A.~Razavian, H.~Azizpour, J.~Sullivan, and S.~Carlsson.
\newblock {CNN Features off-the-shelf: an Astounding Baseline for Recognition}.
\newblock {\em arXiv preprint}, 1403.6382v3, 2014.

\bibitem{romera2018tits}
E.~Romera, J.M. Alvarez, L.M. Bergasa, and R.~Arroyo.
\newblock Erfnet: Efficient residual factorized convnet for real-time semantic
  segmentation.
\newblock {\em IEEE Trans.~on Intelligent Transportation Systems (ITS)},
  19(1):263--272, 2018.

\bibitem{tuba2017wseas}
E.~Tuba, R.~Jovanovic, and M.~Tuba.
\newblock Plant diseases detection based on color features and kapur’s
  method.
\newblock {\em World Scientific and Engineering Academy and Society (WSEAS)
  Trans. Inf. Sci. Appl.}, 14:31--39, 2017.

\bibitem{vysotska2019rsswsabstract}
O.~Vysotska, H.~Kuhlmann, and C.~Stachniss.
\newblock {UAVs Towards Sustainable Crop Production}.
\newblock In {\em Workshop at Robotics: Science and Systems}, 2019.

\bibitem{weyler2021ral}
J.~Weyler, A.~Milioto, T.~Falck, J.~Behley, and C.~Stachniss.
\newblock {Joint Plant Instance Detection and Leaf Count Estimation for
  In-Field Plant Phenotyping}.
\newblock {\em IEEE Robotics and Automation Letters (RA-L)}, 6(2):3599--3606,
  2021.

\bibitem{weyler2022ral}
J.~Weyler, J.~Quakernack, P.~Lottes, J.~Behley, and C.~Stachniss.
\newblock {Joint Plant and Leaf Instance Segmentation on Field-Scale UAV
  Imagery}.
\newblock {\em IEEE Robotics and Automation Letters (RA-L)}, 7(2):3787--3794,
  2022.

\bibitem{weyler2022wacv}
J.~Weyler, F.~Magistri, P.~Seitz, J.~Behley, and C.~Stachniss.
\newblock In-field phenotyping based on crop leaf and plant instance
  segmentation.
\newblock In {\em Proc.~of the IEEE Winter Conf.~on Applications of Computer
  Vision (WACV)}, 2022.

\bibitem{woebbecke1994asabe}
D.M. Woebbecke, G.E. Meyer, K.V. Bargen, and D.A. Mortensen.
\newblock Color indices for weed identification under various soil, residue,
  and lighting conditions.
\newblock {\em Transactions of the American Society of Agricultural and
  Biological Engineers (ASABE)}, 38:259--269, 1994.

\bibitem{wu2018cvpr}
Z.~Wu, Y.~Xiong, S.~Yu, and D.~Lin.
\newblock Unsupervised feature learning via non-parametric instance-level
  discrimination.
\newblock In {\em Proc.~of the IEEE/CVF Conf.~on Computer Vision and Pattern
  Recognition (CVPR)}, 2018.

\bibitem{yoo2016eccv}
D.~Yoo, N.~Kim, S.~Park, A.S. Paek, and I.S. Kweon.
\newblock Pixel-level domain transfer.
\newblock In {\em Proc.~of the Europ.~Conf.~on Computer Vision (ECCV)}, 2016.

\bibitem{you2020computersinagriculture}
J.~You, W.~Liu, and J.~Lee.
\newblock A dnn-based semantic segmentation for detecting weed and crop.
\newblock {\em Computers and Electronics in Agriculture}, 178:105750, 2020.

\bibitem{zbontar2021icml}
J.~Zbontar, L.~Jing, I.~Misra, Y.~LeCun, and S.~Deny.
\newblock Barlow twins: Self-supervised learning via redundancy reduction.
\newblock In {\em Proc.~of the Intl.~Conf.~on Machine Learning (ICML)}, 2021.

\bibitem{zhang2018iclr}
H.~Zhang, M.~Cisse, Y.N. Dauphin, and D.~Lopez-Paz.
\newblock mixup: Beyond empirical risk minimization.
\newblock In {\em Proc.~of the Int.~Conf.~on Learning Representations (ICLR)},
  2018.

\bibitem{zhong2020aaai}
Z.~Zhong, L.~Zheng, G.~Kang, S.~Li, and Y.~Yang.
\newblock Random erasing data augmentation.
\newblock In {\em Proc.~of the Conference on Advancements of Artificial
  Intelligence (AAAI)}, 2020.

\end{thebibliography}

\end{document}